\documentclass{article}
\PassOptionsToPackage{numbers, compress}{natbib}
\usepackage[main, preprint]{neurips_2026}
\PassOptionsToPackage{final}{changes}
\usepackage[pagebackref=true,colorlinks,linkcolor=blue,citecolor=brown]{hyperref}
\usepackage{etoc}
\usepackage[utf8]{inputenc}
\usepackage[T1]{fontenc}
\usepackage{url,booktabs,amsfonts,nicefrac,microtype,xcolor,amsmath,bm,mathtools,amsthm,graphicx,wrapfig,subcaption,changes}

\newtheorem{lemma}{Lemma}
\newtheorem{proposition}{Proposition}
\newtheorem{remark}{Remark}
\newtheorem{corollary}{Corollary}
\title{Iterative Gumbel Planning for Continuous Control}
\author{
  Shaohuai Liu$^{1,*}$ \quad
  Weirui Ye$^{2,*}$ \quad
  Yilun Du$^{3,\dagger}$ \quad
  Le Xie$^{3,\dagger}$ \\
  $^{1}$Texas A\&M University \quad
  $^{2}$MIT \quad
  $^{3}$Harvard University \\
  \texttt{liushaohuai5@tamu.edu} \\
  $^{*}$Equal contribution. \quad $^{\dagger}$Co-corresponding authors. \\
  \url{https://liushaohuai5.github.io/ez_igp/}
}
\begin{document}
\maketitle
\etocdepthtag.toc{main}
\begin{abstract}
Search-based planning remains a powerful policy improvement paradigm in reinforcement learning, driving milestone successes in discrete domains. However, extending this paradigm to continuous control exposes a structural mismatch: search evaluates a finite set of action candidates, whereas continuous RL typically relies on Gaussian policies. We show that winner-action distillation into a continuous Gaussian induces a \textit{projection error} and a stepwise variance-expanding gradient tendency when the selected action is sufficiently far from the current mean. Empirically, this tendency is associated with elevated terminal variance and a return gap on the evaluated tasks.
To address this, we propose \textbf{Iterative Gumbel Planning (IGP)}. By adopting a per-dimension factorized categorical policy, IGP naturally aligns the policy support with the search candidates, eliminating the projection error and reducing the residual approximation to a bounded discretization error. To bypass the combinatorial explosion of high-dimensional discrete action spaces, IGP pairs Gumbel Top-$K$ candidate sampling with an MPPI-style multi-round iterative search to refine the policy at each state. Empirically, IGP achieves state-of-the-art aggregate performance and sample efficiency on DMControl and HumanoidBench. Furthermore, it produces more stable action sequences with lower temporal variance, demonstrating its practical potential for real-world robotic deployment.
\end{abstract}

\section{Introduction}
\label{sec:intro}

\begin{wrapfigure}{r}{0.5\textwidth}
    \centering
    \vspace{-10pt}
    \includegraphics[width=0.49\textwidth]{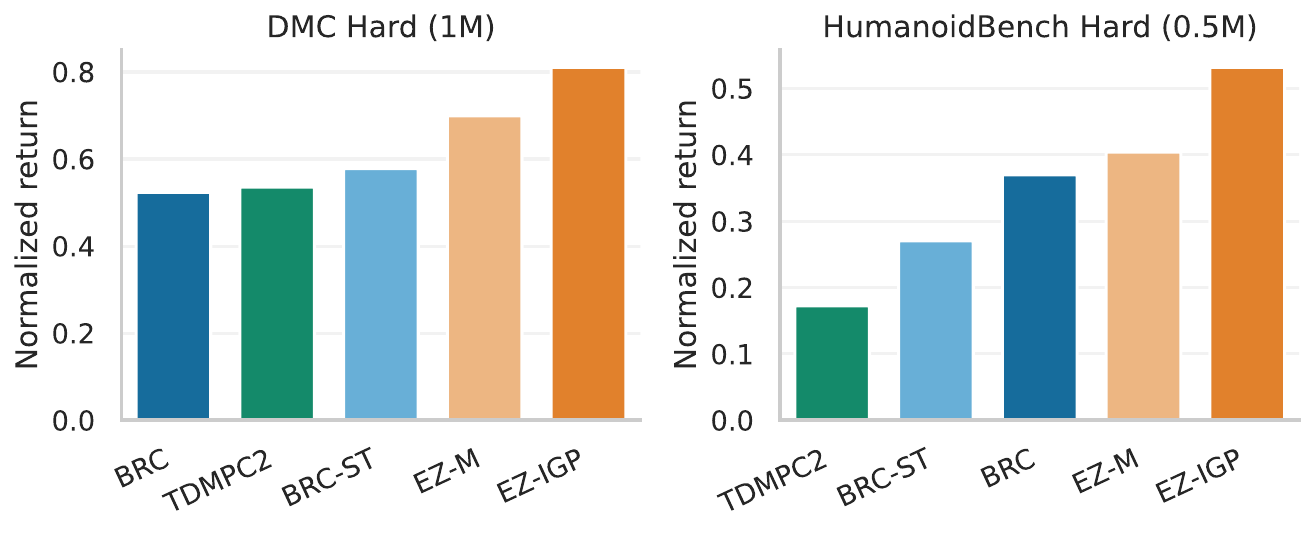}
    \caption{\textbf{Factorized policies achieve SOTA performance.} Across high-dimensional continuous-control benchmarks, our iteratively refined policy consistently attains strong returns. }
    \label{fig:overall_bar}
    \vspace{-10pt}
\end{wrapfigure}

\begin{figure}[h]
\vskip -.7cm
    \centering
    \includegraphics[width=0.95\linewidth]{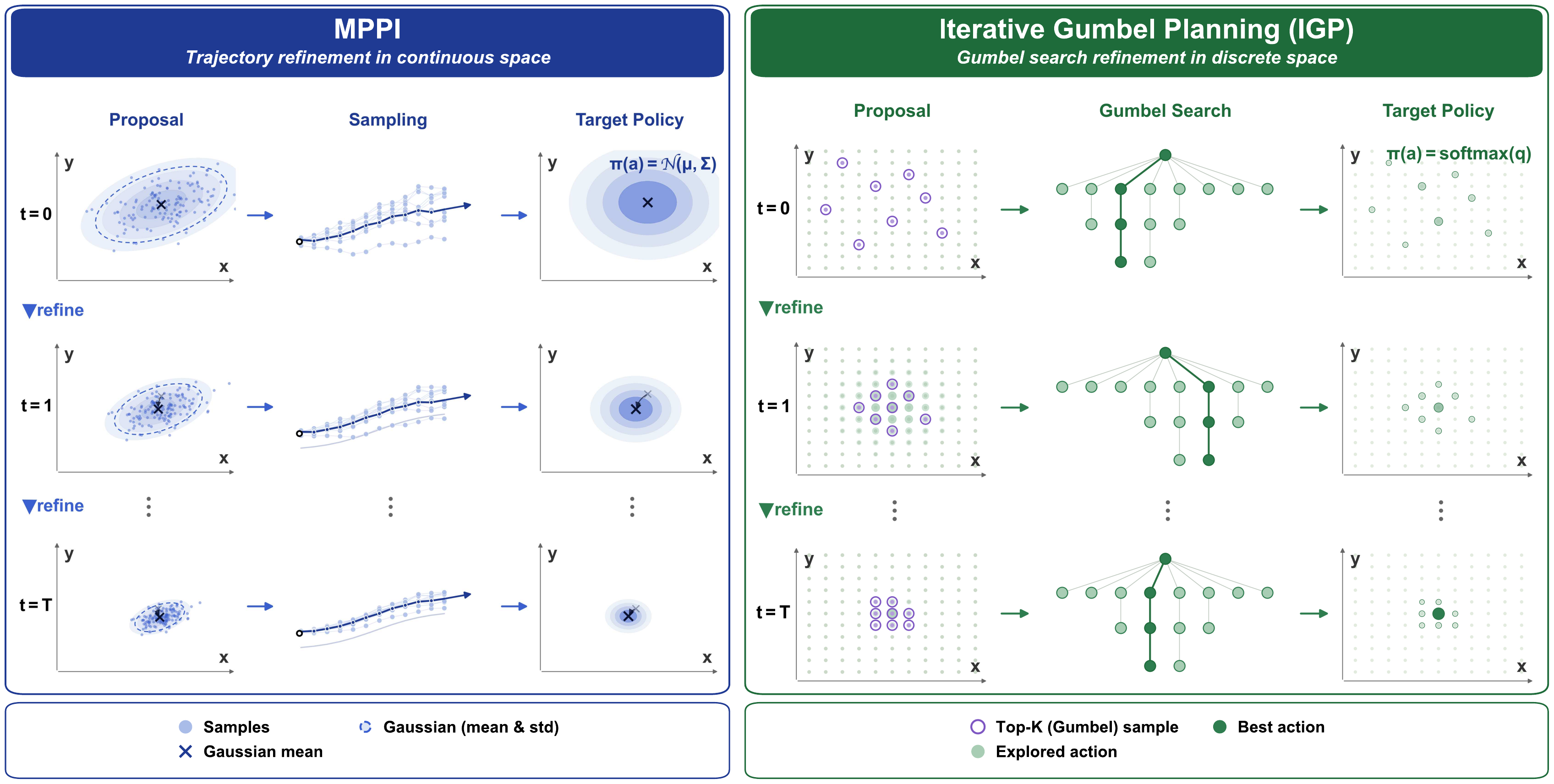}
    \caption{\textbf{Comparison of MPPI and Iterative Gumbel Planning (IGP).} While standard MPPI refines trajectories over an unbounded continuous action space, IGP natively operates in a factorized discrete policy space. By utilizing per-dimension Gumbel Top-$K$ sampling, IGP executes multi-round iterative search to systematically refine a categorical root policy at each state.}
    \vskip -.3cm
    \label{fig:method}
\end{figure}

Continuous control is a key challenge for reinforcement learning, with applications across robotics, control, and beyond. The dominant difficulty lies in credit assignment over long horizons in high-dimensional action spaces. Search-based planning offers a principled response: by combining a learned model with value-guided lookahead, it propagates value estimates over multiple steps and selects actions through explicit comparison rather than single-step bootstrapping. This paradigm has driven major progress on discrete-action benchmarks, with the AlphaGo and MuZero families~\citep{silver2017mastering, schrittwieser2020mastering, hubert2021learning, danihelka2022policy, ye2021mastering} setting the state of the art. Extending it to continuous control is therefore natural, and recent work has begun to explore this direction~\citep{hubert2021learning, wang2024efficientzero}. Yet the continuous setting has not seen comparable dominance, and we argue that the reason lies in a structural mismatch between search and the policy class.

Search is fundamentally a discrete improvement operator: at each state, it evaluates a finite set of candidate actions and returns a categorical distribution over them. In contrast, existing search-based continuous-control methods inherit Gaussian policies from policy-gradient RL~\citep{lillicrap2019continuouscontroldeepreinforcement, haarnoja2018soft} and distill the search output into them via maximum likelihood. Since the selected action is sampled from the Gaussian and rarely coincides with its mean, maximum likelihood not only shifts the mean toward the target but can also encourage larger variance to cover the sampled candidate. We show that winner-action distillation produces a stepwise variance-expanding gradient tendency when the selected action is sufficiently far from the current mean.

To remove this projection error, the policy should be structurally aligned with the discrete search output. A natural solution is to discretize the continuous action space. Although discretization introduces its own approximation gap, we show that this gap is bounded and controllable by the grid resolution, replacing the non-vanishing Gaussian projection error with an explicitly characterized discretization error. The key remaining challenge is the $V^{|A|}$ combinatorial explosion of joint discretization, which has prevented this idea from scaling to high-dimensional control domains.

To resolve this bottleneck, we propose \textbf{Iterative Gumbel Planning (IGP)}. IGP operates on a per-dimension factorized categorical policy, keeping the parameter count linear in $|A|$ while aligning the policy support with the search operator. {Each dimension is sampled without replacement and equally ranked samples are aligned to form $K$ joint candidates. This is a scalable structured approximation, not exact without-replacement sampling from the exponentially large joint factorized distribution.} Crucially, IGP treats search not as a single-pass procedure, but as an MPPI-style multi-round refinement process that repeatedly improves the root policy at the same state.

We implement IGP on EfficientZero-Multitask (EZ-M)~\citep{liu2026scaling} and evaluate it on DMControl~\citep{tassa2018deepmind} and HumanoidBench~\citep{sferrazza2024humanoidbench}, covering high-dimensional locomotion and whole-body humanoid control. IGP achieves state-of-the-art performance, improves sample efficiency over strong model-based baselines, and produces action sequences with lower temporal variance at evaluation, which is relevant to real-world robotic deployment. \textbf{Our contributions are}: (1) We characterize the variance-expanding gradient tendency induced by winner-action Gaussian distillation and contrast its projection error with the bounded, controllable error of discretization; (2) We propose Iterative Gumbel Planning (IGP), which performs MPPI-style multi-round refinement over a factorized categorical policy via Gumbel Top-$K$ sampling, bypassing Gaussian projection; (3) IGP achieves state-of-the-art sample efficiency on DMControl and HumanoidBench, yielding significantly lower temporal action variance and smoother control behaviors at evaluation.

\section{Related Work}

\textbf{Discrete policies for continuous control.}
Discrete action representations have long been explored to enable value-based learning and stabilize continuous control. To address the combinatorial explosion in high-dimensional action spaces, prior work often relies on per-dimension factorization~\citep{tavakoli2018action, tang2020discretizing} or hierarchical action selection~\citep{mao2020poly, seo2024coarse}. More recently, action discretization has also been adopted in vision-language-action (VLA) models~\citep{brohan2022rt, zitkovich2023rt, kim2024openvla, liu2026oat}, mainly due to the autoregressive format of pretrained transformers. In contrast, our work studies discretization from the perspective of search-based RL optimization, aiming to align the policy support with finite-support search targets in high-dimensional continuous control.

\textbf{Search-based policy improvement.}
Search-based methods treat lookahead planning as a policy improvement operator and have driven major progress in discrete-action domains~\citep{silver2017mastering, schrittwieser2020mastering, ye2021mastering, danihelka2022policy}. Existing adaptations to continuous control typically sample candidates from a Gaussian proposal and perform search over this continuous subset~\citep{hubert2021learning, wang2024efficientzero}. However, distilling the resulting finite-support search target back into a continuous Gaussian introduces projection errors and variance inflation, a structural limitation that we formally analyze in \S\ref{sec:analysis_method}.

\textbf{Iterative refinement in model-based planning.}
Model-based planning methods improve decisions through iterative refinement. MPPI~\citep{williams2017information}, TD-MPC2~\citep{hansen2023td}, and related trajectory optimization methods such as CEM repeatedly sample, evaluate, and update continuous action proposals. These approaches naturally operate on continuous distributions, whereas our work translates iterative refinement to a discrete, factorized policy space, enabling search-style improvement without Gaussian projection.
\section{Preliminaries}

We build IGP on MuZero-style search-based policy improvement. We briefly review two relevant components: Sampled MuZero, which extends MuZero to large action spaces through sampling, and Gumbel MuZero, which formulates search as a finite-budget policy-improvement operator.

\textbf{Sampled MuZero with Gaussian proposals.}
Sampled MuZero~\citep{hubert2021learning} extends MuZero to large or continuous action spaces by sampling a small candidate set from a proposal distribution $\beta_\theta$, typically a Gaussian policy. Given candidates $\mathcal{A}_K(s)=\{a_1,\dots,a_K\}\sim \beta_\theta(\cdot\mid s)$, MCTS is applied only over this finite set, producing an improved search policy $\pi_{\mathrm{imp}}$ summarized by visit counts, with $a_{\mathrm{sel}}=\arg\max_k \pi_{\mathrm{imp}}(a_k\mid s)$. This finite-support target is distilled back into the proposal policy, either through a cross-entropy/KL loss restricted to $\mathcal{A}_K(s)$ or by maximizing $\log\beta_\theta(a_{\mathrm{sel}}\mid s)$.

\textbf{Policy improvement with Gumbel search.}
Gumbel MuZero~\citep{danihelka2022policy} casts search as a policy-improvement operator under a finite simulation budget. Given a prior $\beta_\theta(\cdot\mid s)$, it perturbs the prior logits with i.i.d.\ Gumbel noise and applies sequential halving over the selected candidates, allocating simulations to actions that are both likely under the prior and promising under value-guided search. Under suitable assumptions, such as reliable value estimates and correct backups, the resulting search policy $\pi_{\mathrm{imp}}(\cdot\mid s)$ improves over the prior and is used as the training target for $\beta_\theta$. We adopt this construction for the inner search step throughout this paper.
\section{Analysis and Methodology}
\label{sec:analysis_method}

\begin{wrapfigure}{r}{0.6\textwidth}
    \centering
    \vspace{-10pt}
    \includegraphics[width=0.6\textwidth]{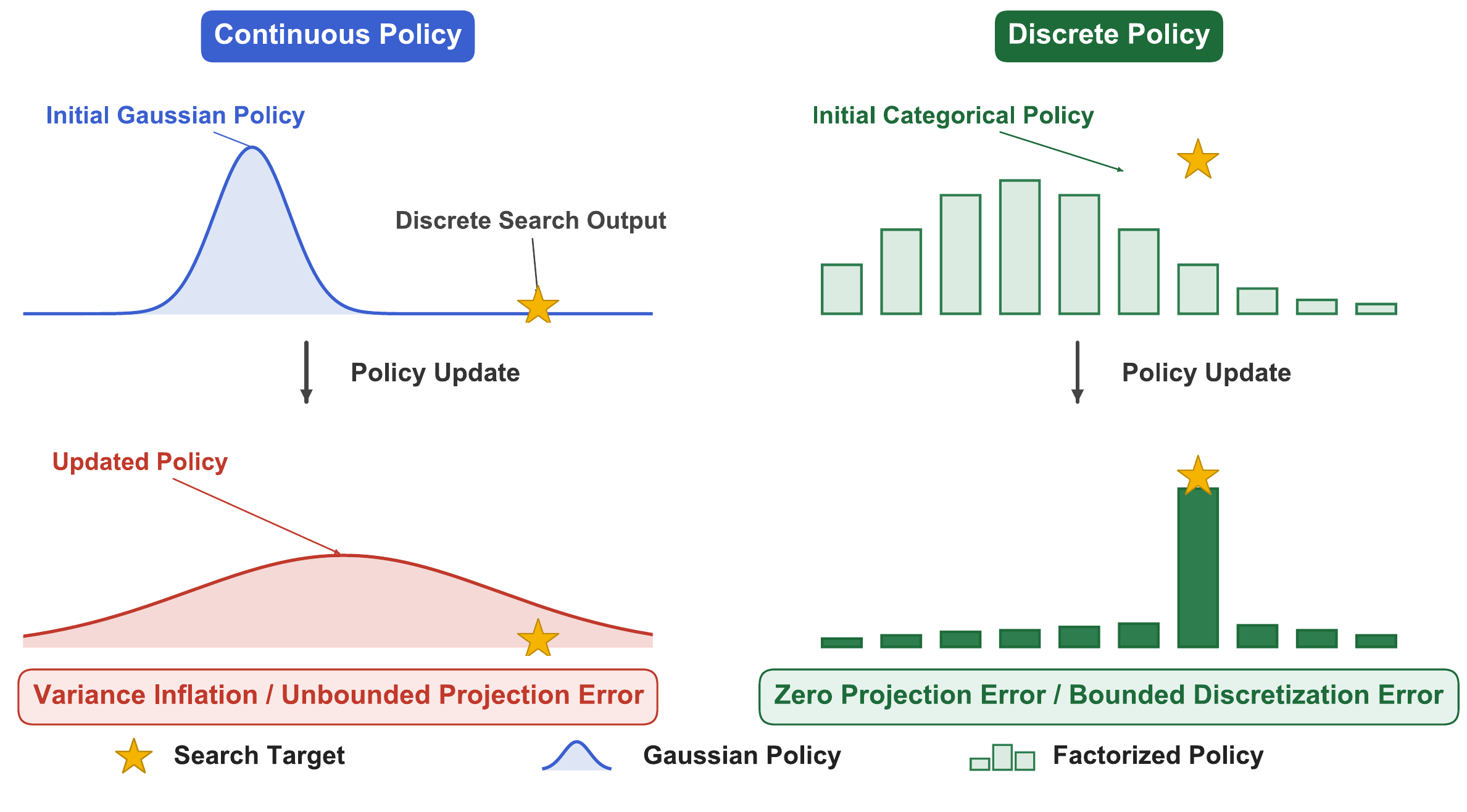}
    \caption{\textbf{Error trade-off in search-guided policy improvement.} (\textit{Left}) Gaussian distillation inherently inflates policy variance due to an unbounded projection error. (\textit{Right}) Our factorized discrete policy naturally aligns with the search support for exact distillation, replacing this unbounded error with a bounded discretization error.}
    \label{fig:theory_gap}
    \vspace{-10pt}
\end{wrapfigure}

In this section, we present the structural analysis and algorithmic design of our approach. As conceptually illustrated in \textbf{Figure~\ref{fig:theory_gap}}, {we analyze a winner-action Gaussian distillation objective that exhibits a variance-expanding gradient tendency under the stated local assumptions, and contrast its continuous projection with the bounded discretization error of a support-aligned categorical policy} (\S\ref{sec:gauss_failure}). To resolve this mismatch, we introduce Iterative Gumbel Planning (IGP), which natively aligns the policy support with the discrete search candidates (\S\ref{sec:igp}), trading the continuous projection error for a rigorously bounded discretization error (\S\ref{sec:tradeoff}).

\subsection{Why Gaussian policies fail under search-based distillation}
\label{sec:gauss_failure}

Consider a Gaussian policy with diagonal covariance $\beta_\theta(\cdot\mid s)=\mathcal{N}(\mu,\mathrm{diag}(\sigma^2))$ at a fixed state $s$. Action samples are drawn via reparameterization,
$
    a_k=\mu+\sigma\odot\epsilon_k,\quad \epsilon_k\overset{i.i.d.}{\sim}\mathcal{N}(0, I),\quad k=1,\ldots,K,
$
and the winner action is the candidate maximizing the search-estimated value $\hat Q$:
$
    a_{\mathrm{sel}}\in\mathop{\arg\max}_{k\in\{1,\ldots,K\}}\hat Q(s,a_k).
$
Following the winner-based maximum-likelihood loss in Gumbel MuZero~\citep{danihelka2022policy}, the policy update maximizes the log-likelihood at the winner:
\begin{equation}
    \mathcal{L}(\mu,\sigma;s)=-\log\beta_\theta(a_{\mathrm{sel}}\mid s).
    \label{eq:winner_mle_loss}
\end{equation}
Distilling a single winner into the Gaussian via~\eqref{eq:winner_mle_loss} is itself a projection step: a finite-support search target is fit by a continuous density. {The following result characterizes the expected directional pressure on the scale parameter under a local first-order model.}

\begin{proposition}[Directional inflation of Gaussian policy variance]
\label{prop:gaussian_var_inflation}
Assume the action value admits a local first-order approximation around $\mu$,
\[
\hat Q(s,a)\approx \hat Q(s,\mu)+g^\top(a-\mu)+o(\|a-\mu\|),\quad g\neq 0.
\]
Let $w=\sigma\odot g$, and let the winner $a_{\mathrm{sel}}$ be treated as a fixed target in the policy update. Then under~\eqref{eq:winner_mle_loss},
\[
\mathbb{E}\!\left[\frac{\partial \mathcal L}{\partial \log\sigma_i}\right]
=
-\frac{w_i^2}{\|w\|^2}\!\left(\mathbb{E}[\epsilon_{(K)}^2]-1\right)
\le 0,
\]
where $\epsilon_{(K)}=\max_{1\le k\le K}\epsilon_k$ with $\epsilon_k\overset{i.i.d.}{\sim}\mathcal{N}(0,1)$. The inequality is strict whenever $w_i\neq 0$ and $K\ge 3$.
\end{proposition}

{\begin{remark}[Weighted search targets]
The same scale tendency is not restricted to a one-hot winner target. For fixed normalized search weights $q_k$ and
\[
\mathcal L_q=-\sum_{k=1}^{K}q_k\log\mathcal N
\!\left(a_k;\mu,\operatorname{diag}(\sigma^2)\right),
\qquad \sum_k q_k=1,
\]
the per-dimension scale gradient is
\[
\frac{\partial\mathcal L_q}{\partial\log\sigma_i}
=1-\sum_{k=1}^{K}q_k\epsilon_{k,i}^{2}.
\]
Thus the update expands variance whenever search places sufficient mass on candidates with above-baseline standardized displacement; winner-action MLE is the one-hot special case. A quadratic variance penalty adds $2\lambda\sigma_i^2$ to this gradient, creating a task- and stage-dependent trade-off between suppressing inflation and preserving proposal diversity.
\end{remark}}

The winner is sampled from $\beta_\theta$ and almost surely does not coincide with $\mu$, so the maximum-likelihood objective raises $\beta_\theta(a_{\mathrm{sel}}\mid s)$ via two mechanisms: shifting $\mu$ toward $a_{\mathrm{sel}}$, and enlarging $\sigma$ to raise the density at $a_{\mathrm{sel}}$. {Proposition~\ref{prop:gaussian_var_inflation} identifies this update-level variance-expanding mechanism, while Figure~\ref{fig:var_comp}(a) shows that it accumulates into a sustained training-time effect under practical finite-budget learning.} The full multi-dimensional proof is in Appendix~\ref{app:proof_of_var_inflation_multi_dim}.

We next show that variance inflation translates into a return gap, even at a converged policy mean.

\begin{proposition}[Inflated variance lowers $Q$]
\label{prop:inf_var_lower_Q}
Suppose the proposal policy is converged in the sense that $\mu=a^*$, and $Q(s,a)$ is locally $m$-strongly concave around $a^*$, a standard regularity assumption near a local optimum. Then
\[
\mathbb{E}_{a\sim\beta_\theta(\cdot\mid s)}[Q(s,a)]
\le
Q(s,a^*) - \frac{m}{2}\sum_i \sigma_i^2.
\]
\end{proposition}

\begin{corollary}[Return degradation under persistent variance inflation]
\label{cor:persistent_variance_return_gap}
If the local concavity condition in Proposition~\ref{prop:inf_var_lower_Q} holds along states in the visitation distribution $d^\pi$, then
\[
J(\pi^*)-J(\pi)
\ge
\frac{m}{2(1-\gamma)}\,\mathbb{E}_{s\sim d^\pi}\!\left[\sum_{i} \sigma_i^2\right].
\]
\end{corollary}

{Under the local concavity assumptions, sustained policy variance induces a return gap proportional to $\sum_i\sigma_i^2$. Figure~\ref{fig:var_comp}(a) empirically demonstrates that the elevated variance persists over the practical training horizon, making this bound operational for the evaluated setting.} Proofs are in Appendix~\ref{app:proof_inf_var_lower_Q_ret}. This motivates aligning the policy support with the search candidates, which we develop next.

\subsection{Iterative Gumbel Planning}
\label{sec:igp}

To avoid the variance inflation analyzed in \S\ref{sec:gauss_failure}, we represent the policy on the same finite support that the search operates on. We instantiate this idea as \textbf{Iterative Gumbel Planning (IGP)}, with four components: a per-dimension factorized categorical policy, Gumbel Top-$K$ candidate sampling, search-output distillation, and multi-round root refinement.

\textbf{Per-dimension factorized categorical policy.}
For a continuous-control task with $|A|$ action dimensions and bounded action space $[-1,1]^{|A|}$, we discretize each dimension into $V$ bins with centers $\{c_v\}_{v=1}^{V}\subset[-1,1]$, giving a grid resolution $\Delta=2/V$. The policy network outputs per-dimension logits $p_\theta(s)\in\mathbb{R}^{|A|\times V}$, where $p_{\theta,j,v}(s)$ is the logit of bin $v$ on dimension $j$. The policy is the product of $|A|$ independent categorical distributions over $V$ bins, and a discrete code $\mathbf z=(z_1,\ldots,z_{|A|})\in\{1,\ldots,V\}^{|A|}$ is decoded into a continuous action via
\begin{equation}
    \mathrm{dec}(\mathbf z) = \frac{2\mathbf z - 1}{V} - 1 \in [-1,1]^{|A|}.
\end{equation}
This factorization keeps the parameter count strictly linear in $|A|\cdot V$, fundamentally avoiding the combinatorial explosion of a joint discrete action space. Per-dimension factorization has been used in prior on-policy work~\citep{tang2020discretizing}. Following standard practice in continuous control RL, where proposal distributions are almost universally parameterized as diagonal Gaussians $\mathcal{N}(\mu,\mathrm{diag}(\sigma^2))$ (e.g., SAC~\citep{haarnoja2018soft}, TD-MPC2~\citep{hansen2023td}, Sampled MuZero~\citep{hubert2021learning}, EfficientZeroV2~\citep{wang2024efficientzero}), our per-dimension factorized categorical policy adopts the \emph{same independence structure} across action dimensions. {Both proposal families therefore omit explicit residual cross-dimensional covariance, although their support and optimization geometry differ. Search evaluates every proposed joint action as a whole, but factorization can still limit proposal quality when strong dependencies must be represented before search.}

\textbf{Gumbel Top-$K$ candidate sampling.}
At each state $s$, IGP constructs $K$ joint candidates using a {per-dimension} Gumbel Top-$K$ trick. For each dimension $j$, we draw i.i.d.\ Gumbel noise $g_{j,v}\sim\mathrm{Gumbel}(0,1)$ and form perturbed scores
\begin{equation}
\tilde p_{j,v}(s)=p_{\theta,j,v}(s) + g_{j,v},
\end{equation}
then take the top-$K$ indices along the bin dimension:
\begin{equation}
\mathcal I_j(s)=\mathrm{TopK}\!\big(\tilde p_{j,1:V}(s),\,K\big)\subset\{1,\ldots,V\},
\qquad |\mathcal I_j(s)|=K.
\end{equation}
We then construct $K$ joint candidates by aligning the $k$-th selected bin across all dimensions:
\begin{equation}
z^{(k)}_j(s)=\mathcal I_j(s)[k],
\qquad
\mathbf z^{(k)}=\big(z^{(k)}_1,\ldots,z^{(k)}_{|A|}\big),
\qquad
a^{(k)}=\mathrm{dec}\big(\mathbf z^{(k)}\big).
\end{equation}
{The Top-$K$ step samples distinct bins without replacement within each dimension, then uses rank alignment to form joint candidates. It is not equivalent to exact Top-$K$ sampling without replacement from the full product distribution; it is a parallelizable structured approximation that avoids enumerating the $V^{|A|}$ joint grid.}

\textbf{Distillation from the search-improved policy.}
Given the candidate set $\{a^{(k)}\}_{k=1}^{K}$, the search procedure produces a distribution $\pi_{\mathrm{imp}}(a^{(k)}\mid s)$ supported on the $K$ candidates, with $\sum_{k=1}^{K}\pi_{\mathrm{imp}}(a^{(k)}\mid s)=1$. We distill $\pi_{\mathrm{imp}}$ into the per-dimension categorical policy by marginalizing it onto each action dimension:
\begin{equation}
t_{j,v}(s)
=
\sum_{k=1}^{K}\pi_{\mathrm{imp}}(a^{(k)}\mid s)\,\mathbb{I}\!\left[z^{(k)}_j=v\right],
\label{eq:marginal_target}
\end{equation}
yielding a valid distribution $t_{j,\cdot}(s)$ on $\{1,\ldots,V\}$ for each $j$. The distillation loss is the sum of per-dimension cross-entropies:
\begin{equation}
\mathcal L_{\mathrm{distill}}(\theta)
=
-\mathbb E_{s\sim\mathcal D}\!\left[
\sum_{j=1}^{|A|}\sum_{v=1}^{V} t_{j,v}(s)\log p_{\theta,j}(v\mid s)
\right].
\label{eq:distill_kl}
\end{equation}
For training stability, we adopt the hard-label special case as default: setting the target to the selected candidate $a^{(k_{\mathrm{sel}})}$ with $k_{\mathrm{sel}}=\arg\max_k \pi_{\mathrm{imp}}(a^{(k)}\mid s)$ reduces~\eqref{eq:marginal_target} to one-hot labels $t_{j,v}(s)=\mathbb{I}[z^{(k_{\mathrm{sel}})}_j=v]$, and~\eqref{eq:distill_kl} becomes a multi-class classification objective across the $|A|$ action dimensions.

\textbf{Multi-round root refinement.}
Because the support of $\pi_{\mathrm{imp}}$ coincides with the policy support, the distillation step in~\eqref{eq:distill_kl} is exact, as we formalize in \S\ref{sec:tradeoff}. This allows IGP to iterate search and refinement at the same state without accumulating projection error. At iteration $t$, MCTS evaluates the current candidate set and returns $\pi_{\mathrm{imp}}^{(t)}(a^{(k)}\mid s)$ together with search scores $q^{(k)}(s)$. These outcomes refine the root logits. {Let $a_t^\star=\arg\max_{a\in\mathcal A_t(s)}\widehat Q(s,a)$. The next round retains this predecessor and samples only the remaining $K-1$ candidates:
\[
\mathcal A_{t+1}(s)=\{a_t^\star\}\cup\widetilde{\mathcal A}_{t+1}(s),
\qquad |\widetilde{\mathcal A}_{t+1}(s)|=K-1.
\]
For a fixed scoring function $\widehat Q$, this gives
$\max_{a\in\mathcal A_{t+1}}\widehat Q(s,a)\ge
\max_{a\in\mathcal A_t}\widehat Q(s,a)$, so refinement cannot discard the best candidate already found. This is only a candidate-retention guarantee, not monotonic policy improvement: changing value estimates, finite search budgets, and later learning updates can still make realized performance non-monotonic.} This is an MPPI-style outer loop in discrete policy space: unlike TD-MPC2~\citep{hansen2023td}, which refines on continuous support, IGP refines the per-dimension categorical logits via Gumbel-sampled discrete candidates.

For the per-step logit update, we use a local kernel around the selected bin index $z^{(k_{\mathrm{sel}})}_j$:
\begin{equation}
\kappa_r\!\left(v, z^{(k_{\mathrm{sel}})}_j\right)
=
\max\!\left(0,\, 1 - \tfrac{|v-z^{(k_{\mathrm{sel}})}_j|}{r+1}\right)
\mathbb{I}\!\left[|v-z^{(k_{\mathrm{sel}})}_j| \le r\right],
\end{equation}
with $r=1$ in practice. After normalizing $\kappa_r$ to sum to one over $v$, we use it as a local bonus $b_{j,v}(s)$, center it under the current categorical $\bar p_{\theta,j}(\cdot\mid s)=\mathrm{softmax}(p_{\theta,j,\cdot}(s))$, and update the logits:
\begin{equation}
\tilde b_{j,v}(s)
=b_{j,v}(s)
-
\sum_{u=1}^{V}\bar p_{\theta,j}(u\mid s)\, b_{j,u}(s),
\qquad
p'_{\theta,j,v}(s)
=
p_{\theta,j,v}(s) + \eta\, \tilde b_{j,v}(s),
\end{equation}
with $\eta=1.0$. This local update spreads probability mass to neighbors of the selected bin, yielding a smooth search-guided refinement rather than an abrupt one-hot replacement.

\subsection{Error trade-off: projection error vs.\ discretization error}
\label{sec:tradeoff}

During the outer learning phase, the target policy $\pi_{\mathrm{imp}}(\cdot\mid s)$ is strictly supported on the finite candidate set evaluated by search. Distilling this discrete target back into the parametric proposal fundamentally acts as a projection onto the policy class:
\begin{equation}
    \beta_{\theta^+}(\cdot\mid s)
    \approx
    \arg\min_{\theta}
    \mathbb{E}_{a\sim \pi_{\mathrm{imp}}(\cdot\mid s)}\!\left[-\log\beta_\theta(a\mid s)\right].
    \label{eq:proj_def}
\end{equation}
For a categorical policy, this objective reduces to cross-entropy on the candidate support; for a Gaussian, it is the negative log-density at the candidates.

\textbf{Gaussian policies: non-vanishing projection error.}
For a Gaussian $\beta_G(\cdot\mid s)=\mathcal{N}(\mu,\mathrm{diag}(\sigma^2))$ and a one-hot target $\pi_{\mathrm{imp}}(a\mid s)=\delta(a-a_{\mathrm{sel}})$, the projection objective
\begin{equation}
    \mathcal{L}_G(s)
    =
    \tfrac{1}{2}\sum_i \tfrac{(a_{\mathrm{sel},i}-\mu_i)^2}{\sigma_i^2}
    +
    \sum_i \log \sigma_i + C
\end{equation}
has the same scale-parameter behavior as the winner-MLE loss in Proposition~\ref{prop:gaussian_var_inflation}: whenever the residual $|a_{\mathrm{sel},i}-\mu_i|>\sigma_i$, gradient descent on $\mathcal{L}_G$ enlarges $\sigma_i$. {In our experiments, the winner target frequently remains far from the current mean and the Gaussian variance stays elevated throughout training, connecting the stepwise mechanism to a sustained long-horizon optimization effect.}

\textbf{Discretized policies: zero projection error on the search support.}
For the per-dimension factorized categorical policy $\beta_D$ of \S\ref{sec:igp}, parameterized by softmax over $\{p_{\theta,j,v}\}$, when search selects a grid action $a_{\mathrm{sel}}$, the one-hot target is achievable in the limit of the policy class:
\begin{equation}
    \inf_{\theta} \mathbb{E}_{a\sim\delta_{a_{\mathrm{sel}}}}\!\left[-\log\beta_D(a\mid s;\theta)\right] = 0,
\end{equation}
where the infimum reflects the standard softmax parameterization (achieved as logits go to infinity). The projection step is therefore exact on the shared support: there is no residual that the policy must absorb as variance.

\textbf{Bounded discretization error.}
Discretization introduces its own approximation gap when the optimal continuous action $a^*(s)$ does not lie on the grid. We bound this gap below.

\begin{lemma}[Discretization error]
\label{lem:discretization_error}
Let $\tilde a^*(s)$ be the nearest grid point to $a^*(s)$ at resolution $\Delta=2/V$, and assume $Q(s,\cdot)$ is $L$-Lipschitz in $a$. Then
\[
\big|Q(s,a^*(s))-Q(s,\tilde a^*(s))\big| \le L\,\tfrac{\sqrt{|A|}}{2}\,\Delta,
\quad
J(\pi^*) - J(\tilde\pi^*) \le \tfrac{L}{1-\gamma}\cdot\tfrac{\sqrt{|A|}}{2}\,\Delta.
\]
\end{lemma}

{For one-hot targets on the grid, the categorical projection error can vanish, whereas Gaussian winner distillation may retain a scale-dependent residual during finite training.} The discretization error is explicitly characterized and shrinks at rate $\Delta=2/V$ as the bin count $V$ increases. The trade-off is favorable: IGP exchanges a continuous-support projection mismatch for a controllable representation error.

\section{Experiments}
\label{sec:exp}

In this section, we evaluate IGP by implementing it on top of EfficientZero-Multitask (EZ-M)~\citep{liu2026scaling}, and refer to the resulting method as EZ-IGP. We continue to adopt the multi-task training setting. Our experiments aim to answer the following questions: (1) How does EZ-IGP perform compared with established baselines? (2) What is the empirical variance distribution when training with a Gaussian policy? (3) What does the learned policy behavior look like in practical applications? (4) How does performance vary with the number of discretization bins? (5) How does performance change with the number of search rounds? We answer these questions in the following subsections, with additional experimental details and training curves provided in Appendix~\ref{sec:app_curves}.

\subsection{Experimental Settings}

\textbf{Benchmarks}. 
We evaluate EZ-IGP on two challenging continuous-control benchmarks: DMControl~\citep{tassa2018deepmind} and HumanoidBench~\citep{sferrazza2024humanoidbench}. Following~\citep{nauman2025bigger,liu2026scaling}, we focus on \textit{DMC-Hard} and \textit{HB-Hard}. \textit{DMC-Hard} includes 7 high-dimensional locomotion tasks: humanoid \texttt{stand}, \texttt{walk}, and \texttt{run}, and dog \texttt{stand}, \texttt{walk}, \texttt{run}, and \texttt{trot}. \textit{HB-Hard} consists of hand-control humanoid tasks, including 9 locomotion tasks as well as \texttt{sit-simple}, \texttt{sit-hard}, \texttt{balance-simple}, \texttt{balance-hard}, and \texttt{reach}. We train all methods for 1 million environment steps; DMControl uses an action repeat of 2, while HumanoidBench uses no action repeat.

\textbf{Baselines}. 
On HumanoidBench, we compare EZ-IGP with representative single-task and multi-task baselines, including TD-MPC2~\citep{hansen2023td}, MH-SAC~\citep{yang2020multi}, BRC~\citep{nauman2025bigger}, and EZ-M~\citep{liu2026scaling}. On DMControl, we compare against TD-MPC2 and EZ-M. EZ-IGP follows the same architectures and training objectives as EZ-M, but replaces its Gaussian policy with our iteratively refined factorized policy. Baseline scores are taken from published results when available, and otherwise from re-evaluations of official implementations. Hyperparameter details are provided in Appendix~\ref{app:hyperparameters}.

\textbf{Evaluation Metrics}. 
We report normalized aggregate scores as described in Appendix~\ref{app:score-norm}, and provide full training curves with original returns in Appendix~\ref{sec:app_curves}.

\subsection{Overall Performance}

\textbf{Main results on DMControl dog and humanoid tasks}. Figure \ref{fig:dmc-unnorm-curves} and Table \ref{tab:dmc_scores_with_igp} show results on the DMControl dog and humanoid locomotion tasks. EZ-IGP consistently outperforms or matches all baselines across the seven environments. The improvement is especially pronounced on the dog tasks. On the four dog tasks, EZ-IGP learns faster and achieves the highest {reported per-seed peak scores}. This indicates that the proposed factorized iterative policy is not only effective for dexterous manipulation-like tasks, but also transfers well to high-dimensional locomotion domains with complex contact dynamics.

\begin{table}[t]
\centering
\scriptsize
\caption{Scores for DMC Hard (1M) tasks, reported as mean $\pm$ standard deviation. Each score is computed by first taking the maximum return within each seed, and then computing the mean and standard deviation across seeds. We \textbf{bold} the best result for each task.}
\label{tab:dmc_scores_with_igp}
\setlength{\tabcolsep}{3pt}
\resizebox{\linewidth}{!}{
\begin{tabular}{lrrrrrr}
\toprule
Task & TD-MPC2 & BRC & BRC-ST & EZ-M-ST & EZ-M & EZ-IGP(ours) \\
\midrule
\texttt{dog-stand} 
& $817.367{\pm}23.354$
& $945.557{\pm}35.713$
& $913.862{\pm}57.743$
& $947.730{\pm}5.045$
& $945.741{\pm}7.064$
& $\mathbf{974.790{\pm}8.358}$ \\

\texttt{dog-walk} 
& $719.833{\pm}68.214$
& $809.929{\pm}147.561$
& $764.241{\pm}97.900$
& $778.395{\pm}110.429$
& $915.123{\pm}6.744$
& $\mathbf{958.983{\pm}5.187}$ \\

\texttt{dog-run} 
& $228.800{\pm}22.439$
& $281.717{\pm}49.678$
& $299.228{\pm}73.987$
& $241.965{\pm}70.275$
& $405.874{\pm}50.808$
& $\mathbf{613.257{\pm}6.546}$ \\

\texttt{dog-trot} 
& $500.033{\pm}16.631$
& $516.288{\pm}130.027$
& $588.662{\pm}95.487$
& $466.416{\pm}122.449$
& $815.958{\pm}76.640$
& $\mathbf{942.814{\pm}3.341}$ \\

\texttt{humanoid-stand} 
& $663.733{\pm}32.515$
& $737.042{\pm}201.735$
& $751.334{\pm}166.116$
& $589.152{\pm}97.187$
& $772.273{\pm}8.943$
& $\mathbf{903.719{\pm}7.734}$ \\

\texttt{humanoid-walk} 
& $628.233{\pm}16.280$
& $314.038{\pm}271.480$
& $569.773{\pm}92.383$
& $531.614{\pm}5.920$
& $779.404{\pm}39.126$
& $\mathbf{911.542{\pm}7.005}$ \\

\texttt{humanoid-run} 
& $190.200{\pm}14.062$
& $62.398{\pm}73.964$
& $161.155{\pm}42.949$
& $153.026{\pm}33.236$
& $259.030{\pm}25.534$
& $\mathbf{369.468{\pm}5.633}$ \\
\bottomrule
\end{tabular}
}
\end{table}

The same pattern appears on the humanoid tasks: EZ-IGP obtains the best {reported peak performance} and generally shows stronger sample efficiency than BRC, BRC-ST, TD-MPC2, and EZ-M, with especially clear gains on \texttt{humanoid-walk} and \texttt{humanoid-run}, where several baselines plateau at lower scores. This suggests that iterative action refinement becomes more beneficial as tasks require more accurate long-horizon coordination.

Taken together, these results show that EZ-IGP provides a simple but effective alternative to continuous Gaussian policy parameterization by factorizing the action space into discrete per-dimension choices and improving candidate actions through iterative refinement. The resulting policy achieves strong sample efficiency, competitive stability, and state-of-the-art or near state-of-the-art performance across both HumanoidBench and DMControl high-dimensional control tasks.

\paragraph{Main results on HumanoidBench.}
Figure~\ref{fig:hb-unnorm-curves} reports learning curves on the high-dimensional HumanoidBench tasks, while full per-task scores and uncertainty are provided in Table~\ref{tab:humanoidbench_hard}. Overall, EZ-IGP achieves strong and consistent performance across a broad set of challenging dexterous-control environments. Compared with EZ-M, EZ-IGP often learns substantially faster and reaches higher final returns, especially on tasks such as \texttt{h1hand-walk}, \texttt{h1hand-run}, \texttt{h1hand-pole}, and \texttt{h1hand-maze}. These tasks require coordinated control over many action dimensions, where Gaussian policy optimization can be sensitive to variance and exploration scale; in contrast, EZ-IGP provides a structured factorized action proposal and further improves selected actions through iterative planning.

The advantage of EZ-IGP is particularly clear in tasks where successful behavior requires precise multi-dimensional coordination rather than simply discovering a coarse locomotion pattern. For example, in \texttt{h1hand-run} and \texttt{h1hand-pole}, EZ-IGP rapidly improves after the early training phase and eventually reaches a much higher score than BRC, BRC-ST, and TD-MPC2. On \texttt{h1hand-reach}, EZ-M achieves the highest score, while EZ-IGP remains stronger than TD-MPC2, BRC, and BRC-ST.

{There are also clear failure cases. EZ-IGP trails the strongest baseline on \texttt{crawl}, \texttt{sit}, and \texttt{balance}. Capacity is a substantial confound: reducing BRC from approximately 1B to 16M parameters lowers its HB-Hard normalized score from $0.576$ to $0.487\pm0.038$, while EZ-IGP reaches $0.743$ at 16M parameters. Nevertheless, BRC-16M remains stronger on the sitting tasks, suggesting a residual task-specific inductive-bias gap. The present comparisons do not isolate finite bin resolution, action factorization, exploration, and objective choice, so we avoid attributing the gap to a single cause.}

\subsection{Policy variance comparisons}
\label{sec:exp:variance}

\begin{figure}[t]
    
    \centering
    
    \begin{minipage}[t]{0.49\textwidth}
        \centering
        \includegraphics[width=\textwidth]{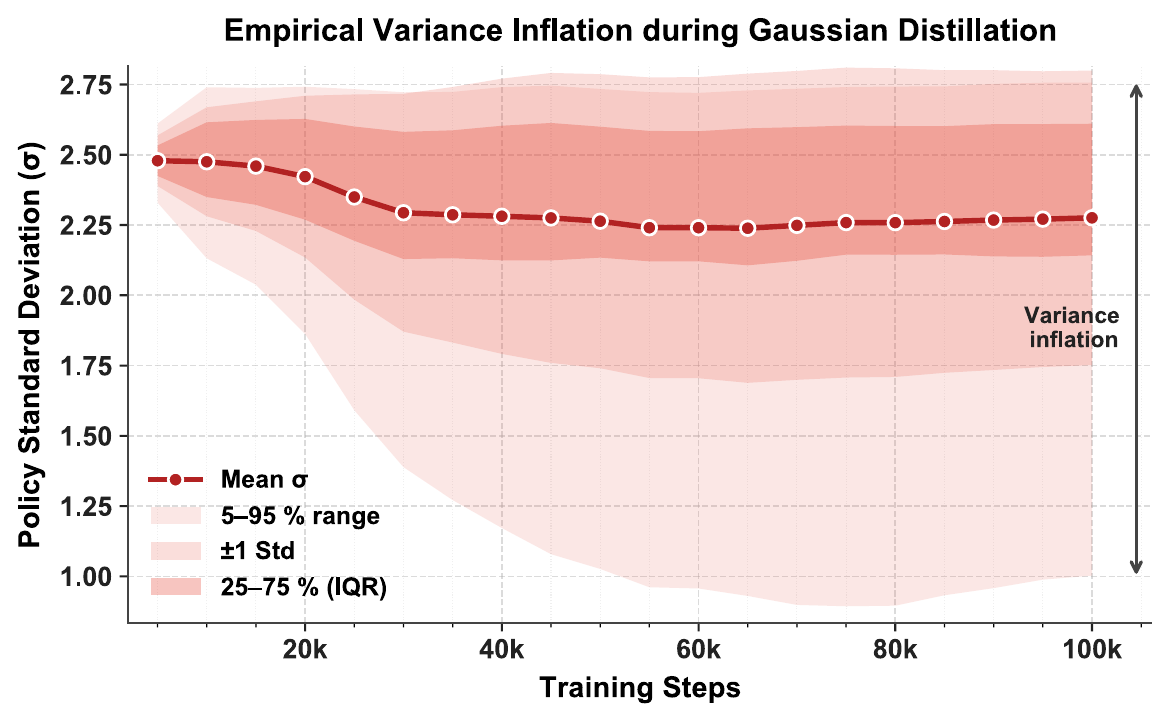}
        \vspace{0.1cm} 
        \centerline{(a) Policy variance distribution in Gaussian distillation.}
    \end{minipage}
    \hfill
    
    \begin{minipage}[t]{0.49\textwidth}
        \centering
        \includegraphics[width=\textwidth]{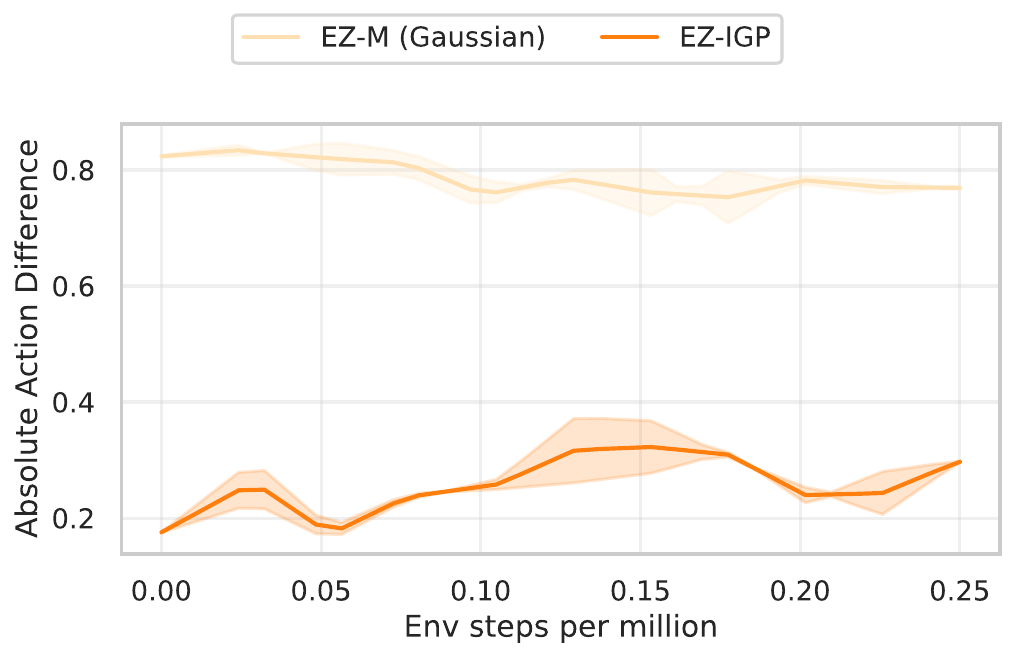}
        \vspace{0.1cm}
        \centerline{(b) Average $|\Delta a|$ across three humanoid tasks.}
    \end{minipage}
    
    \caption{\textbf{Temporal variance comparisons.} \textbf{(a)} {The Gaussian policy standard deviation ($\sigma$) rises early and remains elevated throughout the observed training horizon}; the solid line shows the median and shaded regions show the interquartile and 5--95\% ranges. \textbf{(b)} The Gaussian policy maintains much larger action changes across adjacent timesteps during evaluation, whereas EZ-IGP consistently produces smaller action differences, suggesting smoother and less jittery control behavior.}
    \label{fig:var_comp}
    \vskip -0.1cm
\end{figure}

{Figure~\ref{fig:var_comp}(a) empirically demonstrates the sustained training-time consequence of the update-level mechanism in Proposition~\ref{prop:gaussian_var_inflation}: the standard deviation rises early and remains high throughout the observed training horizon, including its final stage.} The median $\sigma$ stabilizes around a high level, and both the interquartile range and the 5--95\% range suggest that this behavior is not caused by a few outlier dimensions or tasks. Figure~\ref{fig:var_comp}(b) further shows that this elevated variance is reflected in the realized policy behavior: EZ-M with a Gaussian policy produces much larger temporal action differences, while EZ-IGP maintains substantially smoother action sequences. Such action fluctuations can weaken control precision and lead to jittery behaviors in continuous-control tasks. The qualitative comparison videos can be accessed at \href{https://liushaohuai5.github.io/ez_igp/}{the project website}.

\paragraph{{Refinement-matched Gaussian control.}}
{To isolate policy parameterization from multi-round refinement, we train a Gaussian proposal with the same backbone, $K=8$ candidates, $N=16$ simulations per round, two refinement rounds, winner-action target, and unchanged SEM/categorical path. EZ-IGP improves all three DMC humanoid tasks by approximately 139--150 points (Table~\ref{tab:matched_gaussian_refine}). In the matched Gaussian runs, \texttt{dist/policy\_sigma} rises from $1.76\pm0.16$ to an early peak of $3.02\pm0.16$ and remains at $2.11\pm0.04$ over the final 10\% of training. A separate two-seed control that distills a search-score-weighted mixture of all candidates also retains elevated final variance ($2.67\pm0.04$) and underperforms winner-target Gaussian distillation on the same tasks. Thus, neither proposal refinement nor distributing target mass across candidates removes the observed scale pressure.}

\subsection{Ablation Study}

\begin{figure}[t]
    \vskip -.6cm
    \centering
    \includegraphics[width=\linewidth]{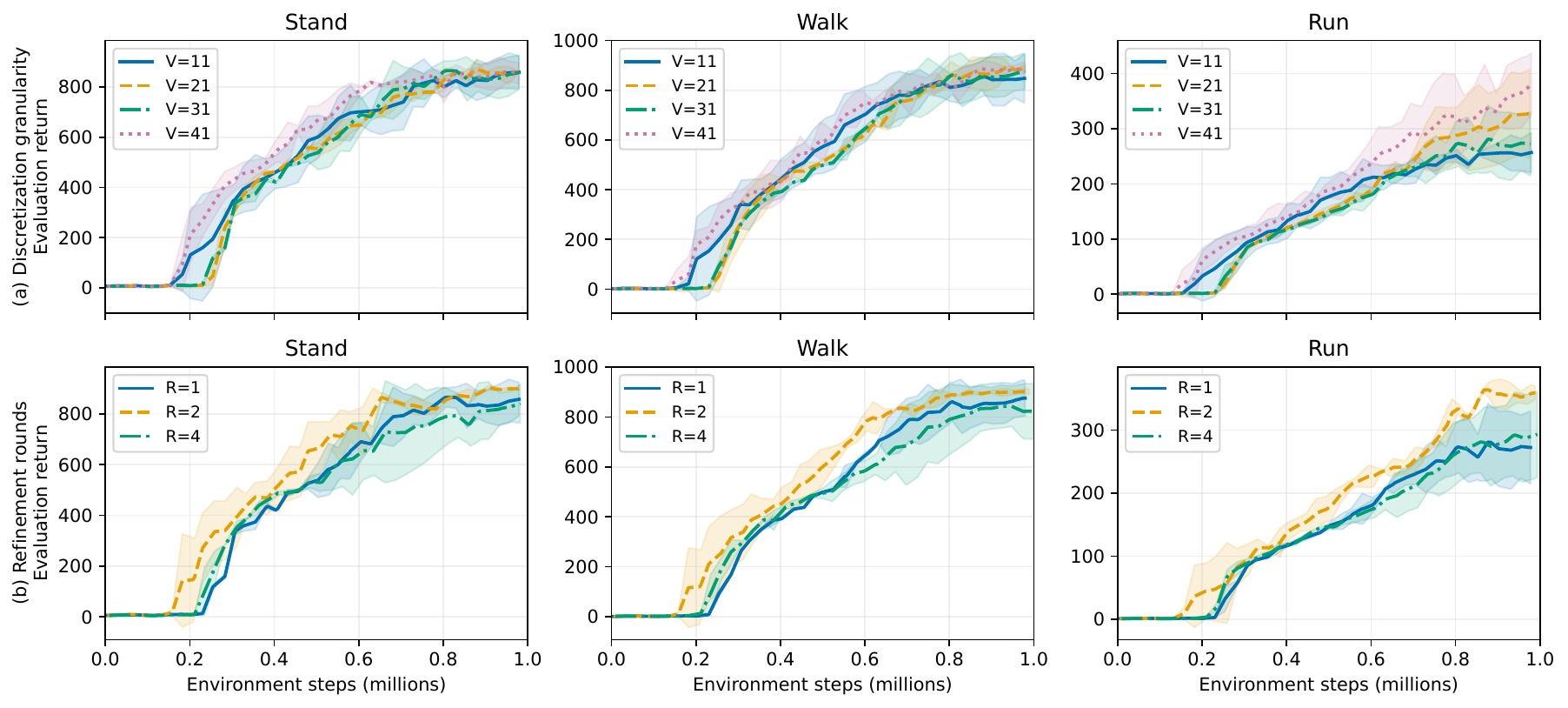}
    
    \caption{\textbf{Ablation studies on three DMC humanoid tasks.} Top: bin count; bottom: refinement rounds. Lines show mean unscaled evaluation returns over two runs, and shading denotes sample standard deviation. Curves are aligned by collected environment steps without endpoint extrapolation. Larger bin counts are not uniformly better; two rounds perform well across these tasks. See Table~\ref{tab:hb_hard_vr_sweep} for HB-Hard.}
    \label{fig:ablations}
    \vskip -.2cm
\end{figure}

\paragraph{Number of factorized bins.}
\added{We investigate the effect of action discretization granularity by varying the number of bins $V$ per action dimension. Figure~\ref{fig:ablations}(a) reports unscaled evaluation returns on three DMC tasks. While different bin counts yield comparable final performance on \texttt{humanoid\_stand} and \texttt{humanoid\_walk}, $V=41$ achieves the highest mean on \texttt{humanoid\_run}, suggesting that finer discretization can be beneficial but does not consistently improve performance. To assess whether this trend generalizes to more challenging control problems, we further evaluate different bin counts across all 14 HumanoidBench-Hard (HB-Hard) tasks, using three seeds per setting, 8 candidates, and 16 simulations per round. As shown in Table~\ref{tab:hb_hard_vr_sweep}, $V=31$ achieves the highest mean performance at $R=2$ after 300k updates, with no further improvement from increasing $V$ to 41. Together, these results suggest that moderate discretization provides sufficient action resolution for effective planning, without requiring an increasingly fine-grained action space. We therefore adopt $V=31$ as a robust choice across benchmarks.}

\paragraph{Number of refinement rounds.}
\added{We next examine the effect of iterative proposal refinement by varying the number of rounds $R$ while fixing $V=31$. Unlike the bin count, which determines the resolution of the action support, $R$ controls how often the search updates its proposal distribution and resamples candidates. With $R=1$, the search performs no proposal refinement, whereas $R=2$ introduces one refinement-and-resampling step. Figure~\ref{fig:ablations}(b) shows that two rounds achieve the highest final mean returns across all three DMC tasks. Increasing to four rounds reduces the terminal means relative to two rounds, although it does not consistently underperform a single round. The broader HB-Hard evaluation in Table~\ref{tab:hb_hard_vr_sweep} reinforces this pattern: increasing $R$ from 1 to 2 improves the mean performance by 0.056, whereas further increasing to four rounds yields no additional gain. These findings indicate that a small number of refinement steps is sufficient to benefit from proposal adaptation, while excessive refinement may concentrate the proposal distribution and limit exploration. We thus select $R=2$ to balance planning effectiveness and computational cost.}

\paragraph{{Sensitivity to the simulation budget.}}
{We vary only the candidate/simulation budget while fixing $V=31$ and two refinement rounds. IGP remains trainable at $2/4$, but performance degrades monotonically as fewer candidates reduce root coverage and fewer simulations weaken candidate ranking (Table~\ref{tab:simulation_budget}). The default $8/16$ setting is therefore an empirical quality--compute trade-off rather than a claim of budget invariance.}

\section{Discussion}
\label{sec:discussion}

This work identifies a structural mismatch between search-based policy improvement and Gaussian policies in continuous control. Since search improves over a finite set of candidate actions, distilling its output into a continuous Gaussian introduces a non-vanishing projection error that can inflate policy variance and degrade returns. IGP addresses this mismatch with a per-dimension factorized categorical policy whose support is aligned with search candidates, replacing Gaussian projection error with bounded discretization error.
Empirically, EZ-IGP achieves strong performance on both DMControl and HumanoidBench, showing that factorized discrete policies can serve as a practical alternative to Gaussian policies in high-dimensional continuous control. \added{The ablations show task-dependent effects of discretization and refinement. Together, DMC and the preliminary HB-Hard sweep support a moderate bin count and two rounds as a practical performance--computation trade-off, rather than establishing a universal optimum or a causal explanation for the effect of additional rounds.}

Several \textbf{limitations} remain. IGP uses fixed uniform discretization and a factorized action representation, which may be suboptimal when action dimensions require different resolutions or exhibit strong dependencies. Its performance also depends on search hyperparameters such as bin count, candidate number, and refinement rounds. Future work could explore adaptive discretization, structured or autoregressive discrete policies, adaptive refinement schedules, and validation on real robots.

\acksection
We thank Jeffrey Polasek and the Texas A\&M University Department of
Electrical and Computer Engineering IT team for their support of the Olympus
computing cluster, which provided computational resources for this work.
Funding is supported in part by Harvard Salata Institute for Climate and Sustainability.
\section*{Disclaimer}
The views expressed in this paper are those of the authors and do not reflect
the views of PJM Interconnection, L.L.C. or its Board of Managers, of which
Le Xie is a member.
\newpage
\bibliography{ref}
\bibliographystyle{unsrt}
\newpage
\appendix
\etocdepthtag.toc{appendix}
\etocsettagdepth{main}{none}
\etocsettagdepth{appendix}{subsection}
\etocsettocstyle{\section*{Appendix Contents}}{}
\tableofcontents
\vspace{1em}
\section{Proofs}
\subsection{Proof of Proposition~\ref{prop:gaussian_var_inflation}}
\label{app:proof_of_var_inflation_multi_dim}
\begin{proof}[Proof: Inflation of Gaussian policy variance (multidimensional perspective)]

Under the local first-order approximation,
\[
\hat Q(s,a)=\hat Q(s,\mu)+g^\top(a-\mu)+o(\|a-\mu\|_2),
\qquad g\neq 0.
\]
Since
\[
a_k=\mu+\sigma\odot \epsilon_k,
\qquad
\epsilon_k\overset{i.i.d.}{\sim}\mathcal N(0,I),
\]
the first-order term becomes
\[
g^\top(a_k-\mu)
=
g^\top(\sigma\odot \epsilon_k)
=
(\sigma\odot g)^\top \epsilon_k.
\]
Define
\[
w\coloneq \sigma\odot g.
\]
Then, up to the local first-order approximation, the winner index is
\[
k^*\in \arg\max_{1\le k\le K} w^\top \epsilon_k.
\]
Let
\[
u\coloneq \frac{w}{\|w\|_2}.
\]
For each sample $\epsilon_k$, decompose it as
\[
\epsilon_k=z_k u+\xi_k,
\]
where
\[
z_k\coloneq u^\top \epsilon_k \sim \mathcal N(0,1),
\qquad
\xi_k \perp u,
\qquad
z_k \text{ is independent of } \xi_k.
\]
Since
\[
w^\top \epsilon_k = \|w\|_2 z_k,
\]
the winner index is equivalently
\[
k^*=\arg\max_{1\le k\le K} z_k.
\]
Hence the selected noise can be written as
\[
\epsilon^* \coloneq \epsilon_{k^*}= z_{(K)}u+\xi^*,
\qquad
z_{(K)}\coloneq \max_{1\le k\le K} z_k,
\]
where $\xi^*$ has the same distribution as the orthogonal component of a standard Gaussian sample and remains independent of $z_{(K)}$.

Now consider the winner-MLE loss
\[
\mathcal L(\mu,\sigma;s)=-\log \pi_{\mu,\sigma}(a^*|s),
\qquad
a^*=\mu+\sigma\odot \epsilon^*,
\]
with $a^*$ treated as a fixed target. For a diagonal Gaussian policy,
\[
\mathcal L
=
\sum_i \left(
\log \sigma_i + \frac{(a_i^*-\mu_i)^2}{2\sigma_i^2}
\right)+C,
\]
where $C$ is constant with respect to $(\mu,\sigma)$. Therefore,
\[
\frac{\partial \mathcal L}{\partial \log \sigma_i}
=
1-\frac{(a_i^*-\mu_i)^2}{\sigma_i^2}
=
1-(\epsilon_i^*)^2.
\]
Taking expectation,
\[
\mathbb E\!\left[\frac{\partial \mathcal L}{\partial \log \sigma_i}\right]
=
1-\mathbb E[(\epsilon_i^*)^2].
\]

Using $\epsilon_i^*=u_i z_{(K)}+\xi_i^*$, we obtain
\[
\mathbb E[(\epsilon_i^*)^2]
=
u_i^2 \mathbb E[z_{(K)}^2]
+
2u_i \mathbb E[z_{(K)}\xi_i^*]
+
\mathbb E[(\xi_i^*)^2].
\]
The cross term vanishes by independence and zero mean of the orthogonal component:
\[
\mathbb E[z_{(K)}\xi_i^*]=0.
\]
Moreover, since a standard Gaussian has unit covariance and $u$ is unit norm,
\[
\mathbb E[(\xi_i^*)^2]=1-u_i^2.
\]
Therefore,
\[
\mathbb E[(\epsilon_i^*)^2]
=
u_i^2 \alpha_K + (1-u_i^2),
\qquad
\alpha_K\coloneq \mathbb E[z_{(K)}^2].
\]
Substituting back,
\[
\mathbb E\!\left[\frac{\partial \mathcal L}{\partial \log \sigma_i}\right]
=
1-\big(u_i^2\alpha_K + 1-u_i^2\big)
=
-(\alpha_K-1)u_i^2.
\]
For $K\ge 3$, we have $\alpha_K>1$, so
\[
\mathbb E\!\left[\frac{\partial \mathcal L}{\partial \log \sigma_i}\right]\le 0,
\]
with strict inequality whenever $u_i\neq 0$, equivalently whenever $(\sigma_i g_i)\neq 0$.
This proves the claim.
\end{proof}

\subsection{Proof of Proposition~\ref{prop:inf_var_lower_Q} and Corollary~\ref{cor:persistent_variance_return_gap}}
\label{app:proof_inf_var_lower_Q_ret}
\begin{proof}[Proof: Inflated variance lowers $Q$ and return]
    For an arbitrary fixed state $s$, assume that $Q(s,a)$ is locally $m$-strongly concave around the optimal action $a^*(s)$, and that there exists a constant $m>0$ such that
\begin{equation}
    Q(s,a)\le Q\bigl(s,a^*(s)\bigr)-\frac{m}{2}\|a-a^*(s)\|^2,
\end{equation}
for all $a$ in a neighborhood of $a^*(s)$.

Consider a Gaussian policy $a \sim \pi(\cdot|s)=\mathcal{N}\bigl(\mu(s),\Sigma(s)\bigr)$, where $\Sigma(s)=\mathrm{diag}(\sigma^2(s))$. Assume the policy is locally converged in the sense that $\mu(s)=a^*(s)$. Then
\begin{equation}
    \mathbb{E}_{a\sim\pi(\cdot|s)}[Q(s,a)]
    \le
    Q\bigl(s,a^*(s)\bigr)
    -\frac{m}{2}\,
    \mathbb{E}_{a\sim\pi(\cdot|s)}\|a-a^*(s)\|^2.
\end{equation}
For the Gaussian policy above,
\begin{equation}
    \mathbb{E}_{a\sim\pi(\cdot|s)}\|a-a^*(s)\|^2
    = \mathrm{tr}\bigl(\Sigma(s)\bigr)
    = \sum_i \sigma_i^2(s),
\end{equation}
and therefore
\begin{equation}
    \mathbb{E}_{a\sim\pi(\cdot|s)}[Q(s,a)]
    \le
    Q\bigl(s,a^*(s)\bigr)
    -\frac{m}{2}\sum_i \sigma_i^2(s).
\end{equation}
Under the assumptions of Proposition~\ref{prop:inf_var_lower_Q}, suppose additionally that for every state $s$ visited by $\pi$, the action $a^*(s)$ maximizes $Q^{\pi^*}(s,a)$, i.e.,
\begin{equation}
    a^*(s)\in \arg\max_a Q^{\pi^*}(s,a),
\end{equation}
and the local quadratic upper bound holds on the support of $\pi(\cdot\mid s)$. Then
\begin{equation}
    J(\pi^*)-J(\pi)
    \ge
    \frac{m}{2(1-\gamma)}
    \mathbb{E}_{s\sim d^\pi}\!\left[\operatorname{tr}(\Sigma(s))\right].
\end{equation}
For diagonal covariance $\Sigma(s)=\operatorname{diag}(\sigma_1^2(s),\dots,\sigma_d^2(s))$, this becomes
\begin{equation}
    J(\pi^*)-J(\pi)
    \ge
    \frac{m}{2(1-\gamma)}
    \mathbb{E}_{s\sim d^\pi}\!\left[\sum_{i=1}^d \sigma_i^2(s)\right].
\end{equation}
\end{proof}

\clearpage
\section{Additional Experimental Results}
\label{app:additional-experiments}

\begin{table}[p]
\centering
\scriptsize
\caption{Scores for HumanoidBench-Hard (1M) tasks, reported as mean $\pm$ standard deviation across three seeds after taking the maximum return within each seed. We \textbf{bold} results above the success score, or the SoTA when all methods are below it.}
\label{tab:humanoidbench_hard}
\resizebox{\linewidth}{!}{
\begin{tabular}{lrrrrrrr}
\toprule
Task & Random & Success & TD-MPC2 & BRC & BRC-ST & EZ-M & EZ-IGP(ours) \\
\midrule
\texttt{h1hand-stand}
& 11.973 & 800.0
& $209.256{\pm}52.474$ & $\mathbf{890.627{\pm}64.554}$ & $344.159{\pm}309.290$ & $\mathbf{914.460{\pm}2.127}$ & $\mathbf{899.706{\pm}11.895}$ \\
\texttt{h1hand-walk}
& 2.505 & 700.0
& $235.559{\pm}93.954$ & $622.684{\pm}139.376$ & $\mathbf{858.652{\pm}106.977}$ & $\mathbf{923.384{\pm}3.694}$ & $\mathbf{{923.752{\pm}3.092}}$ \\
\texttt{h1hand-run}
& 1.927 & 700.0
& $35.943{\pm}3.973$ & $189.786{\pm}25.663$ & $51.847{\pm}13.318$ & $\mathbf{822.343{\pm}43.723}$ & $\mathbf{889.872{\pm}7.244}$ \\
\texttt{h1hand-stair}
& 3.161 & 700.0
& $51.033{\pm}3.398$ & $191.848{\pm}45.344$ & $65.302{\pm}11.846$ & $\mathbf{441.692{\pm}19.294}$ & $381.660{\pm}25.700$ \\
\texttt{h1hand-crawl}
& 278.868 & 800.0
& $\mathbf{898.560{\pm}35.968}$ & $740.718{\pm}59.136$ & $\mathbf{854.395{\pm}19.127}$ & $665.573{\pm}70.116$ & $647.794{\pm}59.109$ \\
\texttt{h1hand-slide}
& 3.142 & 700.0
& $97.656{\pm}16.324$ & $332.524{\pm}58.340$ & $151.583{\pm}39.211$ & $666.507{\pm}65.815$ & $\mathbf{870.521{\pm}46.518}$ \\
\texttt{h1hand-pole}
& 19.721 & 700.0
& $150.233{\pm}21.203$ & $551.138{\pm}104.690$ & $439.295{\pm}28.894$ & $\mathbf{887.945{\pm}25.582}$ & $\mathbf{879.091{\pm}9.683}$ \\
\texttt{h1hand-hurdle}
& 2.371 & 700.0
& $42.457{\pm}6.239$ & $110.333{\pm}11.177$ & $55.887{\pm}9.599$ & $\mathbf{471.050{\pm}38.269}$ & $465.969{\pm}140.063$ \\
\texttt{h1hand-maze}
& 106.233 & 1200.0
& $142.249{\pm}11.927$ & $362.677{\pm}4.216$ & $348.915{\pm}5.388$ & $451.575{\pm}12.825$ & $\mathbf{479.622{\pm}72.949}$ \\
\texttt{h1hand-sit\_simple}
& 10.768 & 750.0
& $675.884{\pm}196.728$ & $\mathbf{935.483{\pm}3.844}$ & $\mathbf{901.682{\pm}25.333}$ & $\mathbf{872.353{\pm}12.648}$ & $\mathbf{814.452{\pm}51.049}$ \\
\texttt{h1hand-sit\_hard}
& 2.477 & 750.0
& $148.216{\pm}36.804$ & $\mathbf{885.141{\pm}33.683}$ & $710.754{\pm}258.261$ & $358.350{\pm}10.554$ & ${426.718{\pm}30.742}$ \\
\texttt{h1hand-balance\_simple}
& 10.170 & 800.0
& $27.412{\pm}3.310$ & $101.305{\pm}8.963$ & $106.005{\pm}6.567$ & $103.400{\pm}12.201$ & $\mathbf{310.167{\pm}17.553}$ \\
\texttt{h1hand-balance\_hard}
& 10.032 & 800.0
& $33.042{\pm}6.165$ & $72.918{\pm}2.266$ & $\mathbf{77.788{\pm}5.622}$ & $74.962{\pm}4.581$ & $70.913{\pm}8.696$ \\
\texttt{h1hand-reach}
& -50.024 & 12000.0
& $3610.454{\pm}617.640$ & $4317.026{\pm}1527.201$ & $4931.460{\pm}597.736$ & $\mathbf{5820.811{\pm}501.354}$ & $5114.852{\pm}40.931$ \\
\bottomrule
\end{tabular}
}
\end{table}

\begin{table}[t]
\centering
\small
\caption{{Refinement-matched comparison on DMC humanoid tasks (mean $\pm$ sample standard deviation over three seeds).}}
\label{tab:matched_gaussian_refine}
\begin{tabular}{lcc}
\toprule
Task & Gaussian + Refine & EZ-IGP \\
\midrule
\texttt{humanoid-stand} & $763.8\pm18.7$ & $\mathbf{903.7\pm7.7}$ \\
\texttt{humanoid-walk} & $761.6\pm35.5$ & $\mathbf{911.5\pm7.0}$ \\
\texttt{humanoid-run} & $230.5\pm37.5$ & $\mathbf{369.5\pm5.6}$ \\
\bottomrule
\end{tabular}
\end{table}

\begin{table}[t]
\centering
\caption{\textbf{Preliminary HB-Hard sensitivity study.} Benchmark normalized scores (mean $\pm$ sample standard deviation over three seeds), evaluated near 300k training updates across 14 tasks. Candidates/simulations are fixed at 8/16 per round; total search computation grows with $R$.}
\label{tab:hb_hard_vr_sweep}
\begin{tabular}{ccc}
\toprule
Bins $V$ & Rounds $R$ & Normalized score \\
\midrule
11 & 2 & $0.622 \pm 0.017$ \\
21 & 2 & $0.663 \pm 0.026$ \\
31 & 2 & $0.743 \pm 0.041$ \\
41 & 2 & $0.690 \pm 0.013$ \\
\midrule
31 & 1 & $0.667 \pm 0.037$ \\
31 & 4 & $0.689 \pm 0.047$ \\
\bottomrule
\end{tabular}
\end{table}

\begin{table}[t]
\centering
\small
\caption{{DMC humanoid performance under strict search budgets (mean $\pm$ sample standard deviation).}}
\label{tab:simulation_budget}
\resizebox{\linewidth}{!}{
\begin{tabular}{ccccc}
\toprule
Candidates / simulations & Stand & Walk & Run & Average \\
\midrule
$2/4$ & $375.193\pm10.490$ & $406.443\pm37.224$ & $108.607\pm0.046$ & $296.748\pm15.889$ \\
$4/8$ & $772.111\pm15.872$ & $790.306\pm47.157$ & $221.021\pm31.538$ & $594.479\pm31.523$ \\
$8/16$ & $903.719\pm7.734$ & $911.542\pm7.005$ & $369.468\pm5.633$ & $728.243\pm1.635$ \\
\bottomrule
\end{tabular}
}
\end{table}

\clearpage
\section{Training Curves}
\label{sec:app_curves}

This section presents the full training dynamics of EZ-IGP and the baselines on the two benchmark suites used in our evaluation: the high-dimensional subset of DMControl (DMC-Hard) and HumanoidBench-Hard. While the per-task scores reported in Tables~\ref{tab:dmc_scores_with_igp} and \ref{tab:humanoidbench_hard} summarize \textit{peak} performance, they do not directly reflect how each method behaves during the course of training. The curves below complement those tables by exposing two additional dimensions of method quality: \textit{learning speed}, i.e., how quickly returns rise as a function of environment interactions, and \textit{learning stability}, i.e., how consistently performance is maintained across seeds and across the training budget.

In all figures, the $y$-axis denotes the unnormalized episodic return and the $x$-axis denotes environment steps in millions. Each curve is aggregated over 5 random seeds; solid lines correspond to the seed mean and shaded regions to 95\% bootstrap confidence intervals. We report \textit{unnormalized} returns here so that readers can directly compare to the raw scores used by prior work and to the per-task random/success thresholds listed in Table~\ref{tab:humanoidbench_hard}.

\begin{figure}[ht]
    \centering
    \includegraphics[width=1.0\linewidth]{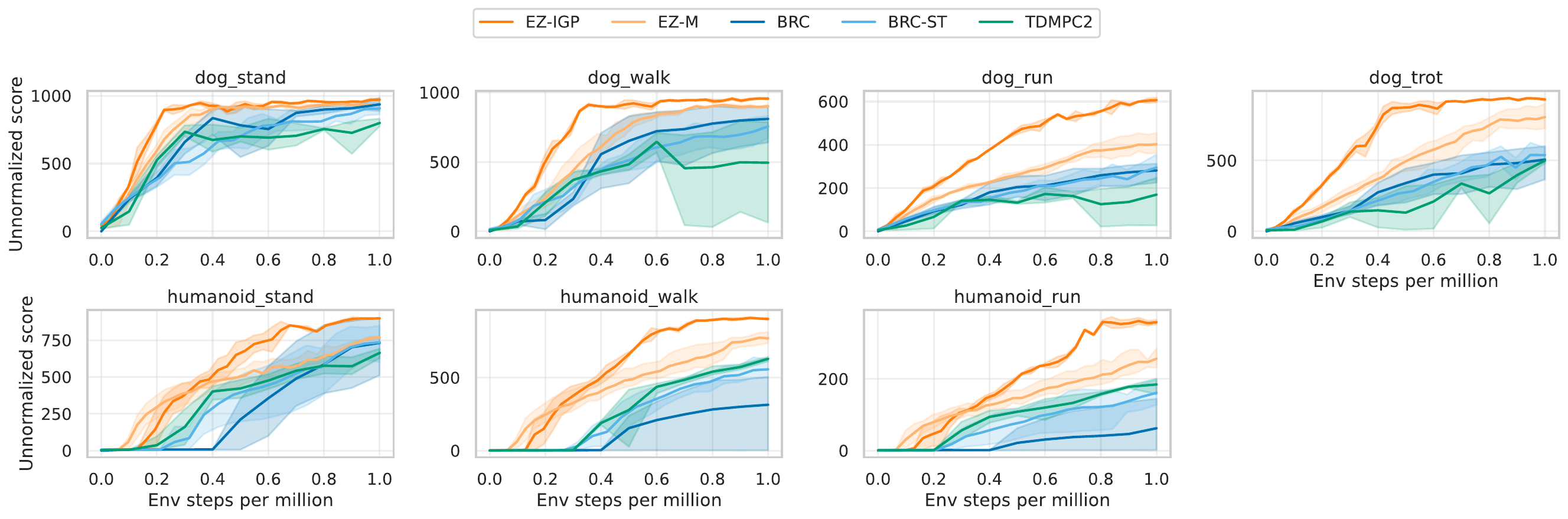}
    \caption{\textbf{Unnormalized curves on DMControl Hard tasks.} The \textit{y}-axis denotes episodic return, and the \textit{x}-axis denotes environment steps (millions). Curves are aggregated over five random seeds, and shaded regions denote 95\% confidence intervals. Per-task scores are listed in Table~\ref{tab:dmc_scores_with_igp}.}
    \label{fig:dmc-unnorm-curves}
\end{figure}

Fig.~\ref{fig:dmc-unnorm-curves} reports the curves on the 7-task DMC-Hard suite, comprising 4 dog tasks (\textit{dog-stand}, \textit{dog-walk}, \textit{dog-run}, \textit{dog-trot}) and 3 humanoid tasks (\textit{humanoid-stand}, \textit{humanoid-walk}, \textit{humanoid-run}). These two embodiment families differ markedly in morphology and physical modeling, so their joint inclusion lets us assess whether the gains from EZ-IGP transfer across embodiments rather than being tied to a specific robot. Reading the curves alongside Table~\ref{tab:dmc_scores_with_igp} also clarifies an effect that peak-score tables can hide: on the more dynamic tasks such as \textit{dog-run} and \textit{dog-trot}, EZ-IGP's advantage opens up early in training and is preserved throughout, indicating better sample efficiency rather than only a higher asymptote.

\begin{figure}[ht]
    \centering
    \includegraphics[width=1.0\linewidth]{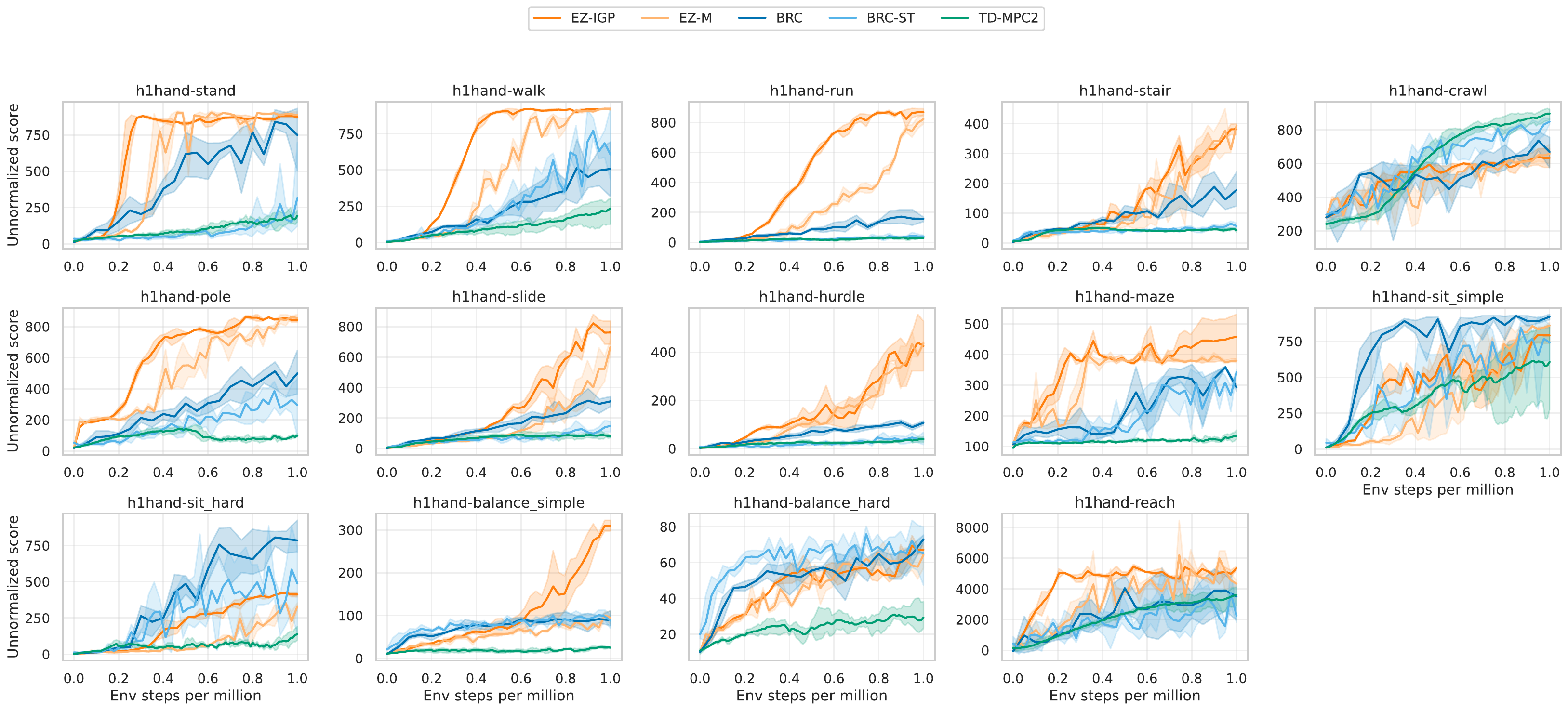}
    \caption{\textbf{Unnormalized curves on the HumanoidBench-Hard benchmark.} The \textit{y}-axis denotes episodic return, and the \textit{x}-axis denotes environment steps (millions). Curves are aggregated over five random seeds, and shaded regions denote 95\% confidence intervals. Per-task random and success thresholds are listed in Table~\ref{tab:humanoidbench_hard}.}
    \label{fig:hb-unnorm-curves}
\end{figure}

Fig.~\ref{fig:hb-unnorm-curves} reports the analogous curves on the 14-task HumanoidBench-Hard suite (H1hand-* with hand control). Compared with DMC-Hard, this setting introduces additional degrees of freedom from finger actuation, more demanding manipulation-style tasks (\textit{sit\_simple}, \textit{sit\_hard}, \textit{balance\_simple}, \textit{balance\_hard}, \textit{reach}), and substantially higher-dimensional observation and action spaces. Beyond confirming the per-task scores in Table~\ref{tab:humanoidbench_hard}, the curves are useful for diagnosing where the gap between EZ-IGP and the strongest baselines opens up during training, and conversely for identifying tasks---such as \textit{h1hand-stair} and \textit{h1hand-sit\_hard}---where the gap is smaller.

\section{Score Normalization}
\label{app:score-norm}

\paragraph{Why normalize.} The benchmarks we consider span tasks whose raw return scales differ substantially---for example, \texttt{h1hand-reach} has a success score of $1.2\times 10^{4}$ while \texttt{h1hand-walk} has a success score of $700$ (Table~\ref{tab:humanoidbench_hard}). An unweighted average of raw returns across such tasks would be dominated by a handful of large-scale tasks and would obscure relative progress on the rest. Following standard practice in the multi-task RL literature, we therefore normalize each task to a comparable scale before aggregating across tasks.

\paragraph{HumanoidBench normalization.} For each HumanoidBench task we map raw episode return to a normalized score using
\begin{equation}
    \text{Return}_\text{norm}=\frac{\text{Return}-\text{Random Score}}{\text{Success Score}-\text{Random Score}},
\end{equation}
where the \emph{Random Score} is the mean episode return obtained by a uniform-random policy on that task and the \emph{Success Score} is the task-specific completion threshold provided by the benchmark authors. Under this definition, a normalized score of $0$ corresponds to the random baseline and a score of $1$ corresponds to nominal task success; values above $1$ are possible and indicate that the agent exceeds the success threshold. The per-task random and success scores are listed in the first two numerical columns of Table~\ref{tab:humanoidbench_hard}.

\paragraph{DMControl normalization.} DMControl returns are by construction bounded in $[0, 1000]$ on each task, so we adopt the simpler convention of dividing the cumulative reward by $1000$ to obtain a normalized score in $[0, 1]$. This avoids introducing a benchmark-specific random-baseline subtraction that is not part of standard DMControl protocol while keeping DMC and HumanoidBench scores on a comparable scale for cross-benchmark comparisons.

\paragraph{Bolding convention and aggregation.} In Table~\ref{tab:dmc_scores_with_igp} we \textbf{bold} the best result per task. In Table~\ref{tab:humanoidbench_hard} we additionally bold any score that surpasses the per-task success threshold, since multiple methods may simultaneously satisfy this criterion; when no method reaches success, we bold only the SoTA. {For both benchmarks and every method, we take the maximum evaluation return within each seed and task, then report the mean and sample standard deviation across seeds. Benchmark aggregates normalize these per-seed task maxima, average uniformly across tasks within each seed, and finally average across seeds.}

\newpage
\section{Hyperparameters and Details}
\label{app:hyperparameters}

This section documents the implementation details of EZ-IGP needed to reproduce our results. All experiments are conducted on a single machine with 4 NVIDIA A100 GPUs. EZ-IGP is built on top of EZ-M and inherits its multi-task training setup, network architecture, and learning objectives; the only methodological change is replacing the Gaussian policy head with the per-dimension factorized categorical policy and the iterative Gumbel planning procedure. The hyperparameters reported here therefore split naturally into two groups: those inherited from EZ-M (optimization, world-model unrolling, replay) and those specific to IGP (bin count, Gumbel candidate count, refinement rounds).

\paragraph{Computational efficiency.}
Our original within-implementation comparison measures 0.368\,s per decision for EZ-IGP (2 rounds, 16 simulations and 8 candidates per round), versus 0.375\,s for EZ-M (1 round, 32 simulations and 16 candidates). These decision-time measurements indicate comparable planning overhead; they do not by themselves establish equal training cost. We separately report the broader baseline comparison below, using a common V100 for decision latency rather than combining timings from different measurement setups.

\paragraph{Baseline decision-time and training-time comparison.}
\added{Table~\ref{tab:baseline_compute} reports a synchronized, warm-started microbenchmark on one NVIDIA V100-SXM2-32GB GPU using HB-Hard observation/action dimensions. Latency includes complete action selection for a single observation, not just a network forward pass. Training times refer to HB-Hard at 1M environment steps per task; BRC timings are read at that budget, rather than at the end of its longer 2M runs. The released TD-MPC2 and BRC-ST curves lack wall-clock timestamps, so we leave these entries unavailable.}

\begin{table}[htbp]
\centering
\caption{\textbf{Baseline compute comparison on HB-Hard.} Decision latency uses the same V100 GPU; training times use the recorded hardware and are not hardware-normalized. BRC-ST latency was not separately measured. Training-time uncertainties are across seeds.}
\label{tab:baseline_compute}
\resizebox{\linewidth}{!}{
\begin{tabular}{lllll}
\toprule
Method & Decision procedure & Latency/action (ms) & Training time at 1M & Training setup \\
\midrule
BRC & Actor only & $0.162\pm0.015$ & $40.58\pm0.03$ h (3 seeds) & 1 RTX 3090/run \\
BRC-ST & Actor only, no task embedding & Not measured & Not timestamped & Separate run/task \\
TD-MPC2 & Full MPPI planning & $47.10\pm0.49$ & Not timestamped & Official curves \\
EZ-IGP & Prediction + two-round Gumbel search & $247.23\pm2.63$ & $24.59\pm0.93$ h (2 seeds) & Distributed A100 setup \\
\bottomrule
\end{tabular}}
\end{table}

\added{BRC is fastest because model-free action selection only evaluates the actor; BRC-ST has the same actor-only structure. TD-MPC2 batches candidate trajectories on the GPU, whereas successive tree expansions depend on preceding search results. Consequently, lower operation counts need not imply lower latency. For context, EfficientZero-v2~\citep{wang2024efficientzero} reports 9,216 latent predictions and $3.6\times10^{10}$ FLOPs for TD-MPC2, versus 32 expansions and $4.7\times10^7$ FLOPs for Gumbel search. EZ-IGP also uses $RN=2\times16=32$ expansions in this comparison; those published FLOP counts are not new measurements of EZ-IGP.}

\added{At the matched interaction budget, EZ-IGP has shorter recorded training wall-clock time than BRC. Its Ray implementation overlaps data collection, reanalysis, and optimization. Because the hardware differs, this is a practical turnaround comparison, not evidence of lower hardware-normalized training compute. BRC-ST additionally requires a separate training run for each task.}

\paragraph{Configuration philosophy.} Unless otherwise specified, we keep the same hyperparameter configuration across \textit{all} tasks within a benchmark. This is a deliberate choice. A central claim of this paper is that the projection-error issue we identify is structural rather than tuning-dependent, so per-task hyperparameter searches would muddy that claim by allowing implicit task-specific adaptation. The only configuration that varies between settings is the number of training steps, which we scale with the size and difficulty of the task set: $2\times 10^{5}$ for DMC-Hard (7 tasks) and $3\times 10^{5}$ for HumanoidBench-Hard (14 tasks, with finger actuation and a higher-dimensional action space).

\paragraph{World-model and optimization settings.} We use Adam with a learning rate of $1\times 10^{-4}$ and weight decay of $2\times 10^{-5}$, which we found to be stable across all task combinations we tried; we did not perform a per-task learning-rate sweep. The discount factor $\gamma=0.99$ and batch size of $512$ are standard for continuous-control RL at this scale. We use a relatively short unroll of $5$ steps and a matching TD horizon of $5$, matching the choice in EfficientZeroV2 and shorter than what fully model-based methods such as DreamerV3 use; this reflects our reliance on MCTS to extend planning depth at decision time rather than at training time.

\begin{table}[t]
\centering
\caption{Key hyperparameters used in HumanoidBench experiments}
\label{tab:key_hyperparams}
\begin{tabular}{ll}
\toprule
\textbf{Hyperparameter} & \textbf{Value} \\
\midrule
Discount factor $\gamma$ & 0.99 \\
Unroll steps & 5 \\
TD steps & 5 \\
Batch size & 512 \\
Replay buffer size & $1\times 10^{6}$\\
Training steps & $2\times 10^{5}$ (DMC) / $3\times 10^{5}$ (HumanoidBench) \\
\midrule
Network hidden dimension & 512 \\
Task embedding dimension & 128 \\
Task embedding usage & Rep / Dyn / Rew / Policy-Value \\
Number of res blocks & 3 \\
\midrule
MCTS simulations & 16 \\
Top actions & 8 \\
Sampled actions & 8 \\
num bins & 31 \\
num refined rounds & 2 \\
\midrule
Optimizer & Adam \\
Learning rate & $1\times 10^{-4}$ \\
Weight decay & $2\times 10^{-5}$ \\
\bottomrule
\end{tabular}
\end{table}

\paragraph{IGP-specific hyperparameters.} Three hyperparameters are introduced by IGP. The number of bins per dimension, $V=31$, controls the discretization resolution and trades off representation precision against the per-dimension softmax size. A finer grid yields a tighter discretization-error bound but enlarges the categorical support; our ablations show that returns improve from $V=11$ up to $V=31$ and then plateau, motivating $V=31$ as a default. The number of Gumbel Top-$K$ candidates per planning call is $K=8$ (\textit{Sampled actions}); together with $8$ \textit{Top actions} for sequential halving and $16$ MCTS simulations per inner search, this fixes the per-round search budget. {We use two refinement rounds as a practical performance--compute trade-off. One round disables iterative refinement, while four rounds provide only a small, higher-variance gain in the broader HB-Hard sweep and double the search expansions; the preferred setting is therefore not universal.}

\paragraph{Task conditioning and shared backbone.} We retain EZ-M's task-conditioning mechanism unchanged: a $128$-dimensional task embedding is injected into all four learned components of the world model---representation, dynamics, reward, and policy-value. Injection at all four sites keeps the embedding small while ensuring that task-specific physics (in dynamics and reward) and task-specific behavior (in representation and policy-value) can both be expressed. The shared backbone uses a hidden dimension of $512$ and 3 residual blocks, also inherited from EZ-M.

\end{document}